\pdfoutput=1

\documentclass[11pt]{article}

\usepackage{acl}

\usepackage{times}
\usepackage{latexsym}

\usepackage[T1]{fontenc}
\usepackage[utf8]{inputenc}

\usepackage{microtype}
\usepackage{adjustbox}
\usepackage{enumitem}
\usepackage{xspace}

\usepackage{subcaption,siunitx,booktabs}

\usepackage{amssymb}%
\usepackage{pifont}%
\newcommand{\cmark}{\textcolor{green}{\ding{51}}}%
\newcommand{\xmark}{\textcolor{red}{\ding{55}}}%
\usepackage{color, colortbl}
\definecolor{LightCyan}{rgb}{0.88,1,1}
\definecolor{Gray}{gray}{0.9}

\newcommand{\demo}{\mathbf{d}}
\newcommand{\obs}{o}
\newcommand{\obss}{\mathbf{\obs}}
\newcommand{\obsspace}{\Omega}
\newcommand{\stat}{s}

\newcommand{\statspace}{\mathcal{S}}
\newcommand{\act}{a}
\newcommand{\acts}{\mathbf{\act}}
\newcommand{\actspace}{\mathcal{A}}
\newcommand{\actstop}{\textsc{stop}\xspace}
\newcommand{\task}{\tau}

\newcommand{\tasks}{\boldsymbol{\task}}
\newcommand{\goal}{g}
\newcommand{\algn}{\alpha}
\newcommand{\algns}{\boldsymbol{\algn}}
\newcommand{\transfn}{T}
\newcommand{\obsfn}{O}
\newcommand{\dataset}{\mathcal{D}}
\newcommand{\datasetann}{\mathcal{D}^{\text{\tiny ann}}}
\newcommand{\pol}{\pi}
\newcommand{\poli}{\pi^{\text{\scriptsize C}}}
\newcommand{\pole}{\pi^{\text{\scriptsize E}}}
\newcommand{\segfn}{\mathrm{seg}}
\newcommand{\seg}{\mathbf{s}}
\newcommand{\params}{\theta}
\newcommand{\qparams}{\eta}
\newcommand{\lik}{\mathcal{L}(\hat{\tasks}, \hat{\algns}, \hat{\theta})}
\newcommand{\likann}{\mathcal{L}^{\text{\scriptsize \normalfont ann}}(\hat{\algns}, \hat{\theta})}

\newcommand{\ourmethod}{(SL)$^3$\xspace}
\newcommand{\ourmethodbf}{\textbf{(SL)}$^{\mathbf{3}}$\xspace}

\SetAlCapFnt{\small}
\SetAlCapNameFnt{\small}

\title{Skill Induction and Planning with Latent Language}

\author{Pratyusha Sharma ~~~ Antonio Torralba ~~~ Jacob Andreas\\
  Massachusetts Institute of Technology \\
  \texttt{\{pratyuss,torralba,jda\}@mit.edu}
  }

\begin{document}
\maketitle
\begin{abstract}
    We present a framework for learning hierarchical policies from demonstrations, using sparse \emph{natural language annotations} to guide the discovery of reusable skills for autonomous decision-making.
    We
    formulate a generative model of action sequences in which goals generate sequences of
    high-level subtask descriptions, and these descriptions generate sequences of
    low-level actions. 
    We describe how to 
    train this model using primarily unannotated demonstrations by \emph{parsing} demonstrations into sequences of named high-level subtasks,
    using only a small number of seed annotations to ground language in action.
    In trained models, natural language commands index a combinatorial library of skills; agents can use these skills to \emph{plan} by generating high-level instruction sequences tailored to novel goals.
    We evaluate this approach in the ALFRED household simulation environment, providing natural language annotations for only 10\% of demonstrations.
    It achieves task completion rates comparable to state-of-the-art models (outperforming several recent methods with access to ground-truth plans during training and evaluation) while providing structured and human-readable high-level plans.\footnote{Code and visualizations:
    \url{https://sites.google.com/view/skill-induction-latent-lang/.}}
\end{abstract}

\section{Introduction}
\label{introduction}

\begin{figure}
\centering
 \includegraphics[width=\columnwidth,clip,trim=0.1in 3.7in 7.1in 1.2in]{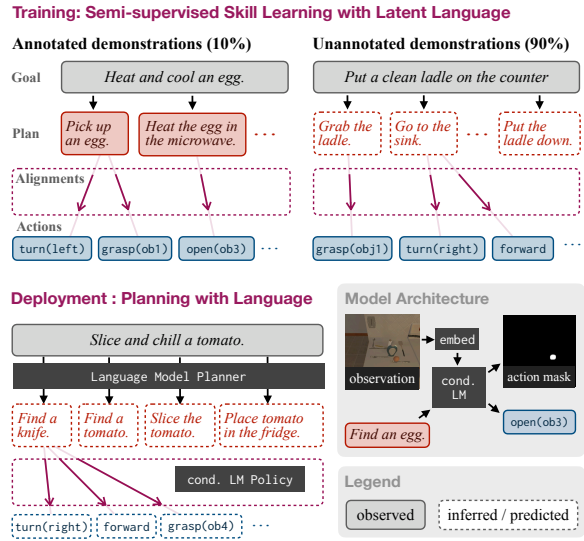}
  \vspace{-1em}
\caption{
Hierarchical imitation learning using weak natural language supervision. During training, a small number of seed annotations are used to automatically segment and label unannotated training demonstrations with natural language descriptions of their high-level structure. When deployed on new tasks, learned policies first generate sequences of natural language subtask descriptions, then modularly translate each description to a sequence of low-level actions.
}
\vspace{-1em}
\label{fig:setup}
\end{figure}

Building autonomous agents that integrate high-level reasoning with low-level
perception and control is a long-standing challenge in artificial intelligence \cite{Fikes1972LearningAE,Newell1973HumanPS,Sacerdoti1973PlanningIA,Brockett1993HybridMF}. \cref{fig:setup} shows an example: to accomplish a task such as \emph{cooking an egg}, an agent must first \emph{find the egg}, then \emph{grasp it}, then \emph{locate a stove or microwave}, at each step reasoning about both these subtasks and complex, unstructured sensor data. \textbf{Hierarchical
planning models} \cite[e.g.][]{options}---which first reason about abstract
states and actions, then ground these in concrete control decisions---play a key
role in most existing agent architectures. But training effective hierarchical
models for general environments and goals remains difficult. Standard techniques
either require detailed formal task specifications, limiting their
applicability in complex and hard-to-formalize environments, or are restricted
to extremely simple high-level actions, limiting their expressive power
\cite{optioncritic,options,maxq,tamp}.

Several recent papers have proposed to overcome these limitations using richer
forms of supervision---especially language---as a scaffold for hierarchical
policy learning.  In \textbf{latent language policies}
\citep[LLPs;][]{andreas-latent-lang},
controllers first map from high-level goals to sequences of natural language
instructions, then use instruction following models to translate those
instructions into actions. But applications of language-based supervision for
long-horizon policy learning have remained quite limited in scope. Current LLP
training approaches 
treat language as a latent variable only during prediction, and 
require fully supervised (and often impractically large) datasets that align
goal specifications with instructions and instructions with low-level actions.
As a result, all existing work on language-based policy learning has focused on
very short time horizons \cite{andreas-latent-lang},
restricted language \cite{Hu2019HierarchicalDM,Jacob2021MultitaskingIS} or
synthetic training data \cite{Shu2018HierarchicalAI,Jiang2019LanguageAA}.

In this paper, we show that it is possible to train language-based hierarchical
policies that outperform state-of-the-art baselines using only minimal natural language supervision.  We introduce a procedure for
\emph{weakly} and \emph{partially supervised} training of LLPs using ungrounded
text corpora, unlabeled demonstrations, and a small set of annotations linking
the two.  To do so, we model \emph{training} demonstrations as generated by latent
high-level plans: we describe a deep, structured latent variable model in which
goals generate subtask descriptions and subtask descriptions generate actions.
We show how to learn in this model by performing inference in the infinite,
combinatorial space of latent plans while using a comparatively small set of annotated demonstrations to seed the learning process.

Using an extremely reduced version of the ALFRED household robotics dataset
\cite{ALFRED20}---with 10\% of labeled training instructions, no alignments during training, and
no instructions at all during evaluation---our approach performs comparably a
state-of-the-art model that makes much stronger dataset-specific assumptions \citep{blukis}, while outperforming several models \citep{hitut, embert, abp} that use more information during both
training \emph{and evaluation}. Our method 
correctly segments and labels subtasks in unlabeled demonstrations,
including subtasks that involve novel compositions of actions and objects.
Additional experiments show that pretraining on large (ungrounded) text corpora
\cite{Raffel2020ExploringTL}
contributes to this success, demonstrating one mechanism by which background
knowledge encoded in language can benefit tasks that
do not involve language as an input or an output.

Indeed, our results show that
relatively little information about language grounding is needed for effective learning of language-based policies---a rich model of natural language text, a large number of demonstrations,
and a small number of annotations suffice for learning compositional libraries of skills and effective policies for deploying them.

\section{Preliminaries}
\label{Preliminaries}

\begin{figure*}
\centering
\includegraphics[height=1.8in,clip,trim=0 5.1in 6.7in 0]{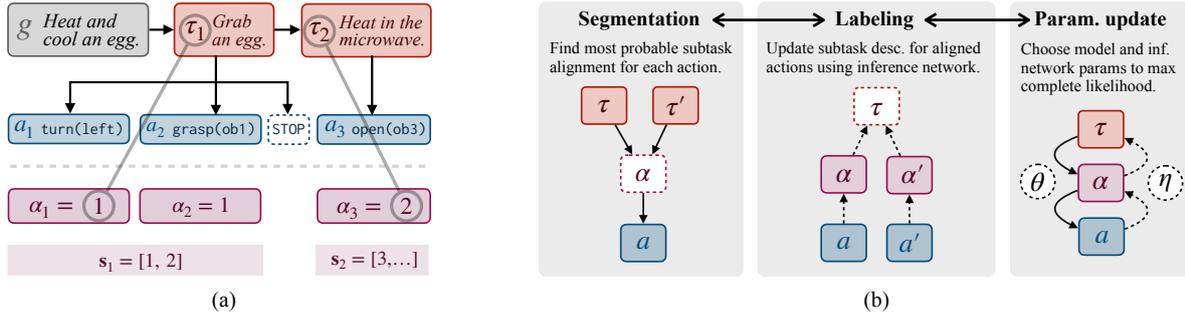}
\hfill
\includegraphics[height=1.8in,clip,trim=4.3in 5.1in 0 0]{figures/variables.pdf}
  \vspace{-1em}
  \caption{(a) 
  When a hierarchical policy is deployed, $\poli$ generates a sequence of subtask specifications, and $\pole$ translates each of these to a low-level action sequence ending in $\actstop$. At training time, this hierarchical structure is not available, and must be inferred to train our model. To do so, we assign each action $\act_i$ an auxiliary \textbf{alignment} variable
  $\alpha_i$ identifying the subtask that produced it. Alignments divide an action sequence into a sequence of
  \textbf{segments} $s$ containing actions aligned to the same subtask. 
  Automatically segmenting training demonstrations makes it possible to
  learn modular, reusable policies for individual subtasks without direct supervision. (b) Overview of the
  proposed learning algorithm \ourmethod, which alternates between segmenting (by aligning) actions
  to fixed subtask specifications; labeling segments given fixed alignments, and
  updating model parameters.}
\label{fig:approach}
\end{figure*}
We consider learning problems in which agents
must perform multi-step tasks (like \emph{cooking an egg}; \cref{fig:setup}) in
interactive environments.  We formalize these problems as undiscounted, episodic, partially
observed Markov decision processes (POMDPs) defined by a tuple $(\statspace,
\actspace, \transfn, \obsspace, \obsfn)$, where $\statspace$ is a set of
\textbf{states}, $\actspace$ is a set of \textbf{actions}, $\transfn: \statspace
\times \actspace  \rightarrow \statspace $ is an (unknown) \textbf{state
transition function}, $\obsspace$ is a set of \textbf{observations}, and $\obsfn
: \statspace \to \obsspace$ is an (unknown) \textbf{observation
function}.\footnote{For notational convenience, we assume without loss of
generality that $\transfn$ and $\obsfn$ are deterministic.} We assume
that observations include a distinguished \textbf{goal specification} $\goal$
that remains constant throughout an episode; given a dataset $\dataset$ of
consisting of \textbf{goals} $\goal$ and \textbf{demonstrations} $\demo$ (i.e.\
$\dataset = \{ (\demo_1, \goal_1), (\demo_2, \goal_2) \ldots \}; \demo =
[(\obs_1, \act_1), (\obs_2, \act_2), \ldots]; \obs \in \obsspace, \act \in
\actspace$), we aim to learn a goal-conditional policy $\pol(\act_t \mid
\acts_{:t-1}, \obss_{:t}, g) = \pol(\act_t \mid \act_1, \ldots, \act_{t-1}, \obs_1, \ldots, \obs_t, g)$ that generalizes demonstrated behaviors to novel goals
and states.

For tasks like the ones depicted in \cref{fig:setup}, this learning problem requires agents to accomplish multiple subgoals (like \emph{finding an
egg} or \emph{operating an appliance}) in a feasible sequence. 
As in past work, we address this challenge by focusing on
\textbf{hierarchical} policy representations that plan over temporal
abstractions of low-level action sequences. We consider a generic class of
hierarchical policies that first predict a sequence of \textbf{subtask
specifications} $\task$ from a distribution $\poli(\task_i \mid \task_{:i-1}, \goal)$ (the
\emph{controller}), then from each $\task$ generate a sequence of actions $\act_1
\ldots \act_n$ from a distribution $\pole(\act_i \mid \acts_{:i-1}, \obss_{:i},
\task)$ (the
\emph{executor}).\footnote{In past work, $\pole$ often conditions on the current observation as well as goal and history of past subtask specifications; we found that this extra information was not needed for the tasks studied here.}
At each timestep, $\pole$ may either generate an action from
$\actspace$; or a special termination signal $\actstop$; after $\actstop$ is
selected, control is returned to $\poli$ and a new $\task$ is generated.  This
process is visualized in \cref{fig:approach}(a). Trajectories generated by
hierarchical policies themselves have hierarchical structure: each subtask
specification $\task$
generates a \textbf{segment} of a trajectory (delimited by a $\actstop$ action)
that accomplishes a specific subgoal.

Training a hierarchical policy requires first defining a space of
subtask specifications $\task$, then parameterizing controller and
executor policies that can generate these specifications appropriately. Most past research has either pre-defined an inventory of
target skills and independently supervised $\poli$ and $\pole$
\cite{options,Kulkarni2016HierarchicalDR,Dayan1992FeudalRL}; or performed
unsupervised discovery of a finite skill inventory using clustering techniques
\cite{maxq,Fox2017MultiLevelDO}. 

Both methods have limitations, and recent work
has explored methods for using richer supervision to guide discovery of skills
that are more robust than human-specified ones and more generalizable than
automatically discovered ones.  
One frequently proposed source of supervision is language: in
\textbf{latent language policies}, $\poli$ is trained to generate goal-relevant
instructions in natural language, $\pole$ is trained to follow instructions, and the space of abstract actions
available for planning is in principle as structured and expressive as language
itself. But current approaches to LLP training remain impractical, requiring
large datasets of independent, fine-grained supervision for $\poli$
and $\pole$. Below, we describe how to overcome this limitation, and instead
learn from large collections of unlabeled demonstrations augmented with only a
small amount of natural language supervision.

\section{Approach}
\label{Approach}

\paragraph{Overview}

We train hierarchical policies on unannotated action sequences by inferring latent natural language descriptions of the subtasks they accomplish (\cref{fig:approach}(b)). We present a learning algorithm that jointly partitions these action sequences into smaller segments exhibiting reusable, task-general skills, labels each segment with a description, trains $\poli$ to generate subtask descriptions from goals, and $\pole$ to generate actions from subtask descriptions.

Formally,
we assume access to two kinds of training data: a large collection of
\textbf{unannotated demonstrations} $\dataset = \{(\demo_1, \goal_1), (\demo_2,
\goal_2), \ldots\}$ and a smaller collection of \textbf{annotated
demonstrations} $\datasetann = \{(\demo_1, \goal_1, \tasks_1), (\demo_2,
\goal_2, \tasks_2), \ldots \}$ where each $\tasks$ consists of a sequence of
natural language \textbf{instructions} $[\task_{1}, \task_{2}, \ldots]$
corresponding to the subtask sequence that should be generated by
$\poli$.  We assume that even annotated trajectories leave much of the
structure depicted in \cref{fig:approach}(a) unspecified, containing no
explicit segmentations or $\actstop$ markers. (The number of instructions $|\tasks|$ will in general be smaller than the number of actions $|\demo|$.) Training $\pole$ requires
inferring the correspondence between actions and annotations on $\datasetann$ while inferring annotations themselves on $\dataset$.

\paragraph{Training objective}
To begin, it will be convenient to have an explicit expression for the probability of a demonstration given a policy $(\poli, \pole)$.
To do so, we first observe that the hierarchical generation procedure depicted in \cref{fig:approach}(a) produces a latent \emph{alignment} between each action and the subtask $\task$ that generated it. We denote these alignments $\algn$, writing $\alpha_i = j$ to indicate that $\act_i$ was generated from $\task_j$.
Because
$\poli$ executes subtasks in sequence, alignments are
\emph{monotonic}, satisfying $\algn_i = \algn_{i-1}$ or $\algn_i =
\algn_{i-1} + 1$.  Let $\segfn(\algns)$ denote the \textbf{segmentation}
associated with $\algns$, the sequence of sequences of action indices $[[i :
\algn_i = 1], [i : \algn_i = 2], \ldots]$ aligned to the same instruction (see \cref{fig:approach}(a)).
Then, for a fixed policy and POMDP, we may write the joint probability of a demonstration, goal, annotation,
and alignment as:
\begin{align}
  p(\demo, \goal, &\tasks, \algns) \propto \prod_{\seg \in \segfn(\algns)}
  \bigg[
  \poli(\task_{\seg} \mid \tasks_{_<\seg}, g) \nonumber \\
  &\times \Big( \prod_{i \in 1..|\seg|} \pole(\act_i \mid \acts_{\seg_{:i-1}},
  \obss_{\seg_{:i}}, \task_{\algn_i}) \Big) \nonumber\\
  &\times \pole(\actstop \mid \acts_\seg, \obss_\seg) \bigg] ~ .
  \label{eqn:causallikelihood}
\end{align}
Here $_<\seg$ (in a slight abuse of notation) denotes all segments preceding $\seg$, and
$\seg_i$ is the index of the $i$th action in $\seg$.
The constant of proportionality in \cref{eqn:causallikelihood} depends only on terms involving $\transfn(\stat' \mid \stat, \act)$, $\obsfn(\obs \mid \stat)$ and $p(\goal)$,
all independent of $\poli$ or $\pole$; \cref{eqn:causallikelihood} thus describes the component of the data likelihood under the agent's control \cite{ziebart2013principle}. %

With this definition, and given
$\dataset$ and $\datasetann$ as defined above, we may train a latent language policy using partial
natural language annotations via ordinary maximum likelihood estimation,
imputing the missing segmentations and labels in the training set jointly with the parameters of $\poli$ and $\pole$ (which we denote $\params$)
in the combined annotated and unannotated likelihoods:
\begin{align}
  &\argmax_{\hat{\tasks}, \hat{\algns}, \hat{\params}} 
  \lik + \likann \\
  \intertext{where}
  &\lik =
  \sum_{(\demo, \goal) \in \dataset} \log p(\demo, \goal, \hat{\tasks}, \hat{\algns}) \\
  &\likann =
  \sum_{(\demo, \goal, \tasks) \in \datasetann} \log p(\demo, \goal, \tasks, \hat{\algns})
\end{align}
and where we have suppressed the dependence of $p(\demo, \goal, \tasks, \algns)$ on $\hat{\params}$ for clarity.
This objective involves continuous parameters $\hat{\theta}$,
discrete alignments $\hat{\algn}$, and discrete labelings
$\hat{\tasks}$. We optimize it via block coordinate ascent on each of these
components in turn: alternating between re-\textbf{segmenting} demonstrations,
re-\textbf{labeling} those without ground-truth labels, and \textbf{updating
parameters}. The full learning algorithm, which we refer to as \ourmethodbf (\emph{semi-supervised skill learning with latent language}), is shown in \cref{algo}, with each
step of the optimization procedure described in more
detail below.

\subsubsection*{Segmentation: \!$\argmax_{\hat{\algns}} \lik\!+\! \likann$}
The segmentation step associates each low-level action with a high-level subtask by finding the highest scoring alignment sequence $\algns$
for each demonstration in $\dataset$ and $\datasetann$.
While the number of possible alignments for a single demonstration is
exponential in demonstration
length, the assumption that $\pole$ depends only on the current subtask implies the following recurrence relation:
\begin{align}
  &\max_{\algns_{1:n}} p(\demo_{1:n}, \goal, \tasks_{1:m}, \algns_{1:n}) \nonumber \\
  &= \max_{i} ~ \Big(\max_{\algns_{1:i}} p(\demo_{1:i}, \goal, \tasks_{1:m-1}, \algns_{1:i}
  ) \nonumber \\ 
  &\qquad \quad \times p(\demo_{i+1:n}, \goal, \task_{m}, \algns_{i+1:n}=m)\Big)
  \label{eqn:dpinvariant}
 \end{align}
This means that the highest-scoring segmentation can be computed by an
algorithm that recursively identifies the highest-scoring alignment to each
prefix of the instruction sequence at each action (\cref{algo2}), a process requiring
$O(|\demo||\tasks|)$ space and $O(|\demo|^2 |\tasks|)$ time. The
structure of this dynamic program is identical to the forward algorithm for
hidden semi-Markov models (HSMMs), which are widely used in NLP for tasks like
language generation and word alignment \cite{Wiseman2018LearningNT}. Indeed, \cref{algo2} can be derived immediately from
\cref{eqn:causallikelihood} by interpreting $p(\demo, \goal, \tasks, \algns)$ as the output distribution
for an HSMM in which emissions are actions, hidden states are alignments, the
emission distribution is $\pole$ and the transition distribution is the
deterministic distribution with $p(\alpha+1 \mid \alpha) = 1$.

This segmentation procedure does not produce meaningful subtask boundaries until an initial executor policy has been trained. Thus, during the first iteration of training,
we estimate a segmentation by by fitting a 3-state hidden Markov model to training action sequences using the Baum--Welch algorithm \cite{baum}, and mark state transitions as segment boundaries.
Details about the initialization step may be found in \cref{app:init}.

\begin{algorithm}[t!]
\small
\SetAlgoLined
\textbf{Input}: Unannotated demonstrations $\dataset = \{(\demo_1, \goal_1), (\demo_2, \goal_2), \ldots\}$; \hspace{-5em} \\ Annotated demonstrations $\datasetann = \{(\demo_1, \goal_1, \tasks_1), (\demo_2, \goal_2, \tasks_2), \ldots \}\hspace{-1.5em}$ \\[1em]

\textbf{Output}: Inferred alignments $\hat{\algns}$, labels $\hat{\tasks}$, and parameters $\params$ for $\poli$ and $\pole$. \\[1em]

// \textbf{Initialization}\\[0.5em]

 Initialize policy parameters $\theta$ using a pretrained language model \cite{Raffel2020ExploringTL}. \\[0.5em]
 
 Initialize inference network parameters $\qparams \gets \argmax_{\hat{\eta}} \sum_{\demo \in \datasetann}
\sum_{\seg, \task} \log
q_\qparams(\task \mid \acts_\seg, \obss_\seg)$. \\[1em]

\For{{\emph{iteration}} $t \gets 1 \ldots T$}{
\strut\\[-0.5em]

// \textbf{Segmentation}\\
// Infer alignments between actions and subtasks. \\
\eIf{$t=1$}{Initialize $\hat{\algns}$ using the Baum--Welch algorithm \cite{baum}}
{
${\hat{\algns}} \leftarrow \argmax_{\hat{\algns}} \lik + \likann$ [\cref{algo2}].
} \strut \\

// \textbf{Labeling}\\
// Infer subtask labels for unannotated demos $\dataset$.
\\
$\hat{\tasks} \leftarrow  \argmax_{\hat{\tasks}} \lik$ \\[1em]

// \textbf{Parameter Update}\\
// Fit policy and proposal model parameters.\\

$\hat{\theta} \leftarrow \argmax_{\hat{\theta}} \lik + \likann$\\
$\hat{\qparams} \leftarrow \argmax_{\hat{\eta}}\sum_\demo
\sum_{\seg, \task} \log
q_\qparams(\hat{\task} \mid \acts_\seg, \obss_\seg)$
}

\caption{\ourmethod: Semi-Supervised Skill Learning with Latent Language}
\label{algo}
\end{algorithm}

\begin{algorithm}[h!]
\small
\SetAlgoLined
\textbf{Input}: Demonstration $\demo = [(\obs_1, \act_1), \ldots, (\obs_n, \act_n)$; \\
Task specifications $\tasks = [\task_1, \ldots, \task_m]$. \\
Executor $\pole(\acts \mid \obss, \task) = \prod_i \pole(\act_i \mid \acts_{:i-1}, \obss_{:i}, \task)$ \\[1em]

\textbf{Output}: Maximum \emph{a posteriori} alignments $\algns$. \\[1em]

\texttt{scores} $\gets$ an $n \times m$ matrix of zeros \\
// $\texttt{scores}[i,j]$ holds the log-probability of the \\
// highest-scoring sequence whose final action $i$ is \\
// aligned to subtask $j$. \\[1em]

\For{$i \gets 1\ldots n$}{
    \For{$j \gets 1 \ldots |\tasks|$}{
        $\texttt{scores}[i, j] \gets -\infty$ \\
            \For{$k \gets 1 \ldots i-1$}{
                $\texttt{scores}[i, j] \gets \max\big($ \\
                \hspace{1em} $\texttt{scores}[i, j],$ \\ \hspace{1em} $\texttt{scores}[k, j-1]$\\ $\hspace{2em} + \log \pole(\acts_{k+1:i} \mid \obss_{k+1:i}, \tau_j)\big)$
            }
    }
} \strut \\
The optimal alignment sequence may be obtained from \texttt{scores} via back-tracing \cite{rabiner}. \\
\caption{Dynamic program for segmentation}
\label{algo2}
\end{algorithm}

\subsubsection*{Labeling: $\argmax_{\hat{\tasks}} \lik$}

Inference of latent, language-based plan descriptions in unannotated demonstrations involves an intractable search over string-valued $\tasks$. To approximate this search tractably, we used a \textbf{learned, amortized inference procedure} \cite{Wainwright2008GraphicalME,Hoffman2013StochasticVI,Kingma2014AutoEncodingVB} to impute descriptions given fixed segmentations. During each parameter update step (described below), we train an inference model $q_\qparams(\task \mid \acts_{\seg^{(i)}}, \acts_{\seg^{(i+1)}}, \goal)$ to approximate the posterior distribution over descriptions for a given segment given a goal, the segment's actions, and the actions from the subsequent segment.\footnote{In our experiments, conditioning on observations or longer context did not improve the accuracy of this model.} Then, during the labeling step, we label complete demonstrations by choosing the highest-scoring instruction for each trajectory independently:
\begin{multline}
    \hspace{-1em} \argmax_\tasks \log p(\demo, \goal, \tasks, \algns) \approx \\ \Big[ \! \argmax_\task q(\task \,|\, \acts_{\seg^{(i)}}, \acts_{\seg^{(i+1)}}, \goal) \, \Big| \, \seg^{(i)} \!\! \in \!  \segfn(\algns) \Big] \hspace{-.5em}
\end{multline}
Labeling is
performed only for demonstrations in $\dataset$, leaving the labels for
$\datasetann$ fixed
during training.

\subsubsection*{Param update:\! \scalebox{.98}{$\argmax_{\hat{\theta}} \lik \!+\! \likann$}}

This is the simplest of the three update steps: given fixed instructions and
alignments, and $\pole$, $\poli$ parameterized as neural networks, this objective is
differentiable end-to-end. In each iteration, we train these
to convergence (optimization details are described in \cref{Experimental Setup} and \cref{app:model}). During the
parameter update step, we also fit parameters $\qparams$ of the proposal model to maximize the likelihood $\sum_\demo
\sum_{\seg, \task} \log
q_\qparams(\hat{\task} \mid \acts_\seg, \obss_\seg)$ with respect to the current segmentations
$\hat{\seg}$ and labels $\hat{\task}$.

\paragraph{} As goals, subtask indicators, and actions may all be encoded as natural language strings, $\poli$ and $\pole$ may be implemented as conditional language models. As described below, we initialize both policies with models \emph{pretrained} on a large text corpora. 

\section{Experimental Setup}
\label{Experimental Setup}

Our experiments aim to answer two questions. First, does the latent-language policy representation described in \cref{Approach} improve downstream performance on complex tasks? Second, how many natural language annotations are needed to train an effective latent language policy given an initial dataset of unannotated demonstrations?

\paragraph{Environment}
We investigate these questions in the ALFRED environment of \citet{ALFRED20}. ALFRED consists of a set of interactive simulated households containing a total of 120 rooms, accompanied by a dataset of 8,055 expert task demonstrations for an embodied agent annotated with 25,743 English-language instructions. Observations $\obs$ are bitmap images from a forward-facing camera, and actions $a$ are drawn from a set of 12 low-level navigation and manipulation primitives. Manipulation actions (7 of the 12) additionally require predicting a mask over the visual input to select an object for interaction. See \citet{ALFRED20} for details.

While the ALFRED environment is typically used to evaluate \emph{instruction following} models, which map from detailed, step-by-step natural language descriptions to action sequences \cite{ALFRED20,singh2020moca,Corona2021ModularNF}, our experiments focus on an \emph{goal-only} evaluation in which agents are given goals (but not fine-grained instructions) at test time. Several previous studies have also considered goal-only evaluation for ALFRED, but most use extremely fine-grained supervision at \emph{training time}, including full supervision of symbolic plan representations and their alignments to demonstrations \cite{film,hitut}, or derived sub-task segmentations using ALFRED-specific rules \cite{blukis}. In contrast, our approach supports learning from partial, language-based annotations without segmentations or alignments, and this data condition is the main focus of our evaluation.

\paragraph{Modeling details}  $\poli$ and $\pole$ are implemented as sequence-to-sequence transformer networks \citep{vaswani}. $\poli$, which maps from text-based goal specifications to text-based instruction sequences, is initialized with a pre-trained \texttt{T5-small} language model \citep{Raffel2020ExploringTL}. $\pole$, which maps from (textual) instructions and (image-based) observations to (textual) actions and (image-based) object selection masks is also initialized with \texttt{T5-small}; to incorporate visual input, this model first embeds observations using a pretrained \texttt{ResNet18} model \cite{He2016DeepRL} and transforms these linearly to the same dimensionality as the word embedding layer. Details about the architecture of $\poli$ and $\pole$ may be found in \cref{app:model}.

\paragraph{Model variants for exploration} %
In ALFRED, navigation in the goal-only condition requires exploration of the environment, but no exploration is demonstrated in training data, and techniques other than imitation learning are required for this specific skill. To reflect this, we replace all annotations containing detailed navigation instructions \emph{go to the glass on the table to your left} with generic ones \emph{find a glass}. Examples and details of how navigation instructions are modified can be found in \cref{app:nav} and \cref{fig:nav-annotations-modified}.
The ordinary \ourmethod model described above is trained on these abstracted instructions. 

A key advantage of \ourmethod is modularity: individual skills may be independently supervised or re-implemented. To further improve \ourmethod's navigation capabilities, we introduce two model variants in which sub-task specifications beginning \emph{Find\ldots} are executed by a either a planner with ground-truth environment information or a specialized navigation module from the HLSM model \citep{blukis} rather than $\pole$. Outside of navigation, these models preserve the architecture and training procedure of \ourmethod, and are labeled \texttt{\textbf{\ourmethod{}+planner}} and \texttt{\textbf{\ourmethod{}+HLSM}} in experiments below.

\paragraph{Baselines and comparisons}
We compare the performance of \ourmethod to several baselines:

\texttt{\textbf{seq2seq}}: A standard (non-hierarchical) goal-conditioned policy, trained on the $(\goal, \demo)$ pairs in $\dataset \cup \datasetann$ to maximize $\sum_{\acts, \obss, \goal} \log \pi(\acts \mid \obss, \goal)$, with $\pi$ parameterized similar to $\pole$.

\texttt{\textbf{seq2seq2seq}}: A supervised hierarchical policy with the same architectures for $\poli$ and $\pole$ as in \ourmethod, but with $\poli$ trained to generate subtask sequences by maximizing $\sum_{\tasks, \goal} \log \poli(\tasks \mid \goal)$ and $\pole$ trained to maximize $\sum_{\acts, \obss, \tasks, \goal} \log \pole(\acts \mid \obss, \tasks, \goal)$ using only $\datasetann$.
Because $\pole$ maps from complete task sequences to complete low-level action sequences, training of this model involves no explicit alignment or segmentation steps.

\texttt{\textbf{no-pretrain}, \textbf{no-latent}}: Ablations of the full \ourmethod model in which $\poli$ and $\pole$ are, respectively, randomly initialized or updated only on $\likann$ during the parameter update phase.

We additionally contextualize our approach by comparing it to several state-of-the-art models for the \emph{instruction following} task in ALFRED: 
\textbf{\texttt{S+}} \cite{ALFRED20}, \textbf{\texttt{MOCA}} \cite{singh2020moca}, \textbf{\texttt{Modular}} \cite{Corona2021ModularNF}, \textbf{\texttt{HiTUT}} \cite{hitut}, \textbf{\texttt{ABP}} \cite{abp}, \textbf{\texttt{ET}} \cite{et}, \textbf{\texttt{EmBERT}} \cite{embert}, and \textbf{\texttt{FILM}} \cite{film}.
Like \texttt{seq2seq}, these are neural sequence-to-sequence models trained to map instructions to actions; they incorporate several standard modeling improvements from the instruction following literature, including progress monitoring \cite{Ma2019SelfMonitoringNA} and pretrained object recognizers \cite{singh2020moca}. %
Many of these models are trained with stronger supervision than \ourmethod, including instructions and alignments during training, and ground truth instructions \emph{during evaluation}; see \cref{tab:info-diff} for details.

\paragraph{Evaluation}

Following \citet{ALFRED20}, \cref{tab:m1baselines}(a) computes the \emph{online}, \emph{subtask-level} accuracy of each policy, and \cref{tab:m1baselines}(b) computes the end-to-end \emph{success rate} of each policy. See the ALFRED paper for details of these evaluations. 
For data-efficiency experiments involving a large number of policy variants (\cref{tab:ablations}, \cref{fig:supervision}), we instead use an \emph{offline evaluation} in which we measure the fraction of subtasks in which a policy's predicted actions (ignoring object selection masks) \emph{exactly match} the ground truth action sequence.

\begin{table}[t]
\begin{subtable}{1\linewidth}
\caption{\textbf{Online} subtask success rate for \ourmethod and baselines}
\centering
\resizebox{\linewidth}{!}{%
\begin{tabular}{lccccccccc}
\toprule
\textbf{Model} & \bf{\rot{90}{Avg}} &
\rot{90}{\it Clean}                  &
\rot{90}{\it Cool}                   &
\rot{90}{\it Heat}                   &
\rot{90}{\it Pick}                   &
\rot{90}{\it Put}                    &
\rot{90}{\it Slice}                  &
\rot{90}{\it Toggle}                  & 
\rot{90}{\it GoTo}                  \\
\cmidrule{2-10}
\cmidrule{2-10}
\bf{\ourmethod (10\%)}     & 50   &  56 & 75 & 74 & 50 & 48 & 54 & 32 &   13\\
\ourmethod (100\%) & 53 &  68 & 82 & 75 & 50 & 45 & 55 & 32 & 15 \\
\texttt{seq2seq} & 25 &  16 & 33 & 64 & 20 & 15 & 25 & 13 & 14 \\
\texttt{seq2seq2seq} & 39 &  15 & 69 & 58 & 29 & 42 & 50 & 32 & 15\\
\bottomrule
\end{tabular}
}
\end{subtable}

\vspace{1em}

\begin{subtable}{1\linewidth}
\caption{\textbf{End-to-end} task success rates for \ourmethod and other models.}
\centering
\footnotesize
\resizebox{\linewidth}{!}{%
\begin{tabular}{lrllr}
\multicolumn{2}{c}{\bf Goal + partial plan sup.}  && \multicolumn{2}{c}{\bf Extra information} \\
\cmidrule{1-2} \cmidrule{4-5}
Model           & SR && Model           & SR \\
\cmidrule{1-2} \cmidrule{4-5}
\ourmethod (10\%) & \phantom{0}0.0 && 
FILM (Min+21) \nocite{film}     & 20.1 \\
\ourmethod+HLSM (10\%)    & 16.1 && 
\ourmethod+planner (10\%) & 40.4 \\
HLSM (Blukis+21)\nocite{blukis}$^*$   & 17.2 &&
HiTUT (Zhang+21)\nocite{hitut}       & 11.1 \\
\texttt{seq2seq}         & \phantom{0}0.0 && 
EmBERT (Suglia+21)            & \phantom{0}5.7 \\
\texttt{seq2seq2seq}     & \phantom{0}0.0 &&  
ET (Pashevich+21)\nocite{et}             & \phantom{0}7.3 \\
\cmidrule{1-2} 
&&& ABP (Kim+21) \nocite{abp}           & 12.6 \\
&&& \texttt{S+} (Shridhar+20)             & \phantom{0}0.1 \\
&&& MOCA (Singh+21) \nocite{singh2020moca}           & \phantom{0}5.4 \\
\cmidrule{4-5}
\end{tabular}
}
\end{subtable}
\caption{(a) Evaluation of \ourmethod and baselines using the \emph{subtask} evaluation from \citet{ALFRED20}. All models in this section were trained with both goals $g$ and annotated subtask descriptions $\tasks$, but observed only goals during evaluation. %
(b) Evaluation of \ourmethod and concurrent work using the \emph{success rate} evaluation from \citet{ALFRED20}. Models in the left column use only goals and partial subtask descriptions at training time, and only goals at test time. (The HLSM model also uses a rule-based, ALFRED-specific procedure for converting action sequences to high-level plan specifications.) Models on the right use extra information, including ground-truth training segmentations and alignments, and ground-truth test-time plans. {\small *Result of our HLSM reproduction using public code and parameters.}
} 

\label{tab:m1baselines}
\vspace{-1em}
\end{table}

\section{Results}
\label{Evaluation}

\cref{tab:m1baselines} compares \ourmethod with flat and hierarchical imitation learning baselines. The table
includes two versions of the model: a \emph{100\%} model trained with full instruction supervision ($|\dataset| = 0$, $|\datasetann| = 21000$) and a \emph{10\%} model trained with only a small fraction of labeled demonstrations ($|\dataset| = 19000$, $|\datasetann| = 2000$). 
\texttt{seq2seq} and \texttt{seq2seq2seq} models are \emph{always} trained with 100\% of natural language annotations. %
Results are shown in \cref{tab:m1baselines}.
We find:

\begin{figure}
\centering
\vspace{1em}
\includegraphics[width=\columnwidth]{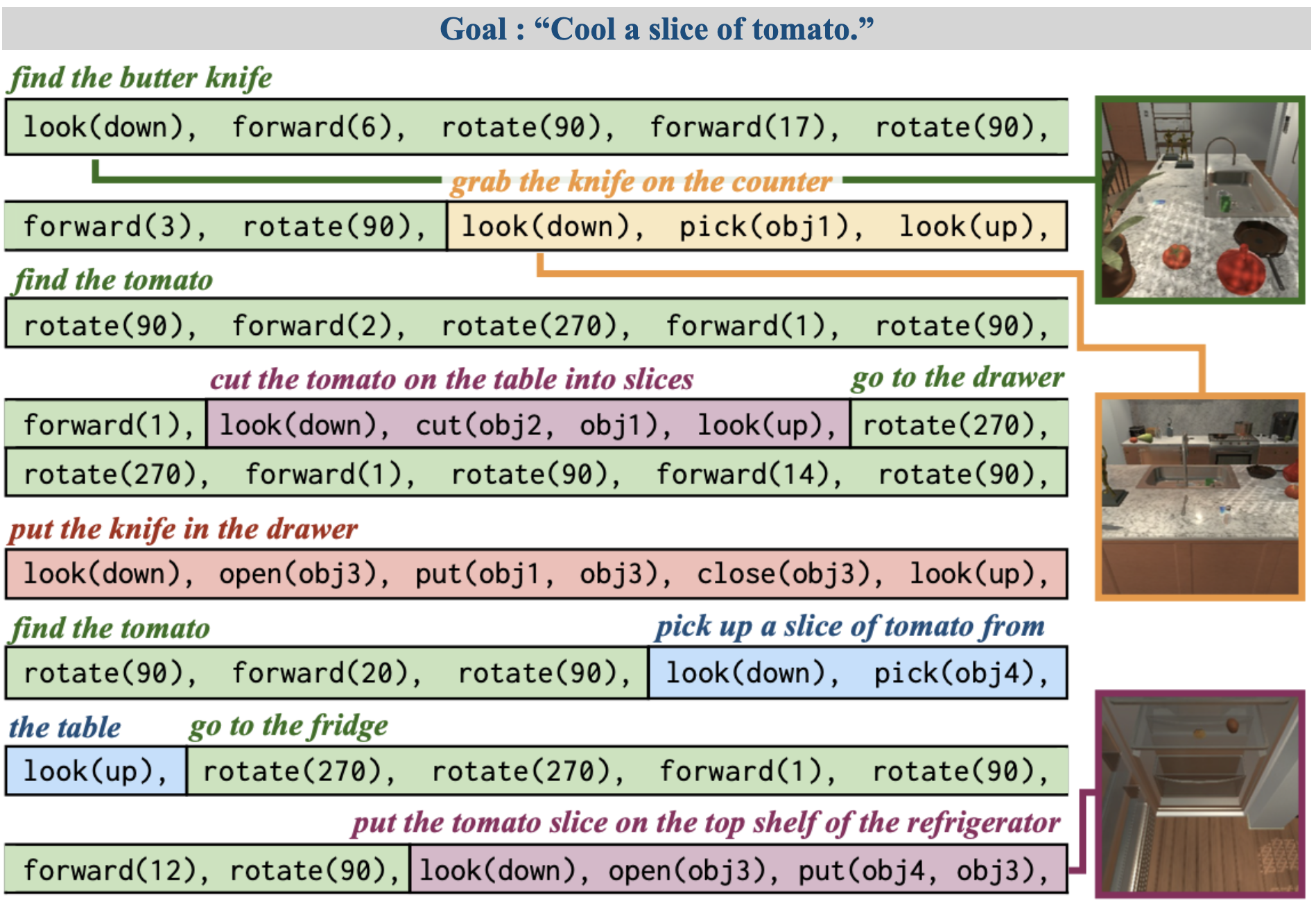}
\vspace{-1.5em}
\caption{Example of an inferred segmentation and labeling for an unannotated trajectory. The trajectory is parsed into a sequence of 10 segments and  $q_\qparams$ assigns high scoring natural-language labels to the segmented actions. These are consistent with the objects, receptacles and sub-tasks. The overall sequence of latent-language skills is a good plan for the high-level goal. }

\label{fig:plans}
\end{figure}

\begin{table}[t]
\centering
\footnotesize
\begin{tabular}{lc}
\toprule
\textbf{Model} & Average \\
\cmidrule{2--2}
\bf \ourmethod (10\%) & 56\\
\ourmethod (100\%) & 58\\
\ourmethod (ground-truth $\algns$) &65\\
\texttt{no-pretrain} & 49\\
\texttt{no-latent} & 52\\
\bottomrule
\end{tabular}
\caption{Ablation experiments. Providing ground-truth alignments at training time improves task completion rates, suggesting potential benefits from an improved alignment procedure. Pretraining and inference of latent task representations contribute 7\% and 4\% respectively to task completion rate with 10\% of annotations.}
\label{tab:ablations}
\end{table}

\textbf{\ourmethod improves on flat policies:}
In both the 10\% and 100\% conditions, it improves over the \emph{subtask} completion rate of the \texttt{seq2seq} (goals-to-actions) model by 25\%. 
When either planner- or mapping-based navigation is used in conjunction with \ourmethod, it achieves \emph{end-to-end} performance comparable to the HLSM method, which relies on similar supervision. Strikingly, it outperforms several recent methods with access to even more detailed information at training or evaluation time.

\begin{figure}[t!]
\centering
\vspace{-.5em}
\includegraphics[width=\linewidth]{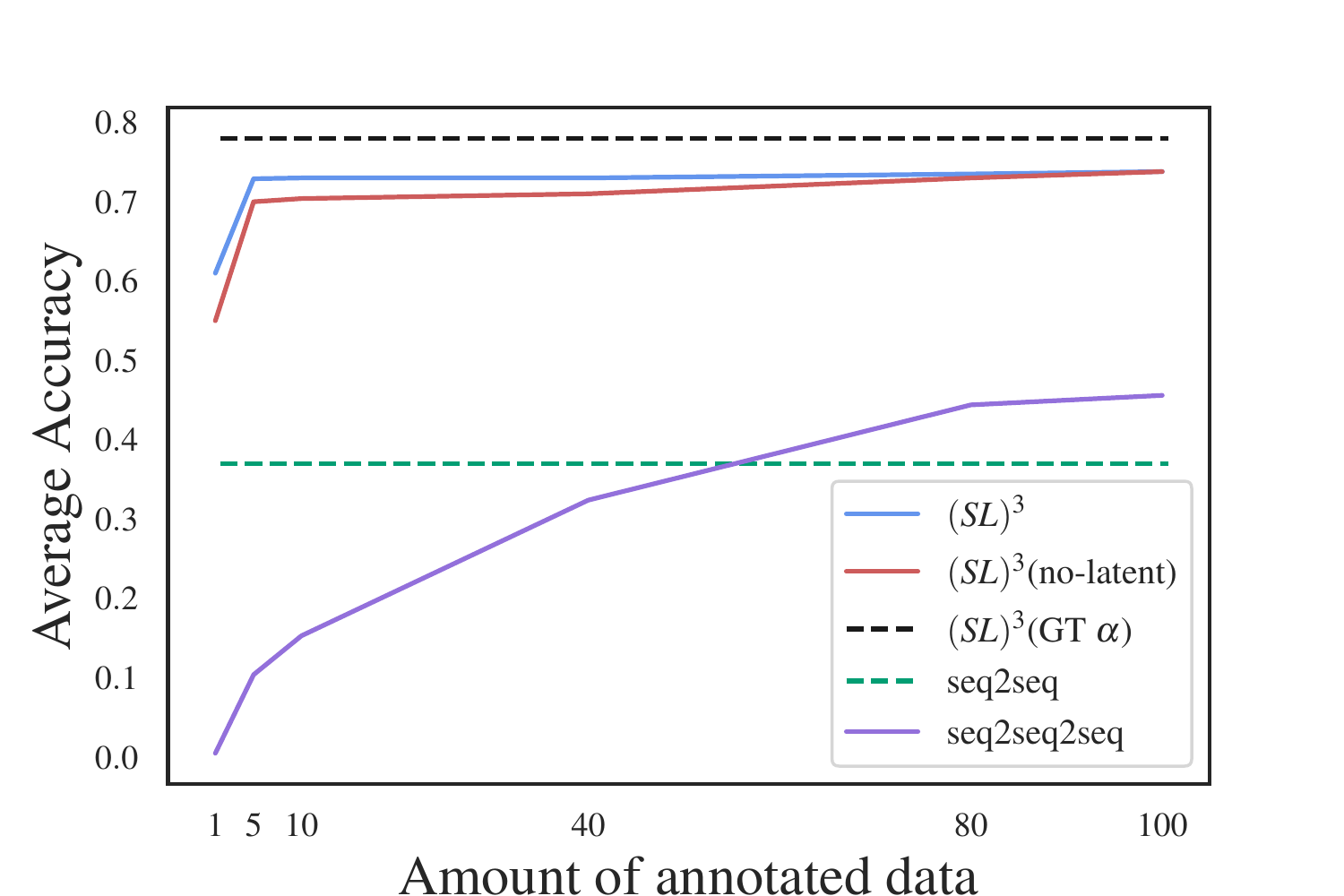}
\vspace{-1em}
\caption{ Offline subtask success rate as a function of
the fraction of annotated examples. Only a small fraction of annotations (5--10\%) are needed for good performance; inference of latent instructions is beneficial
in the low-data regime.}
\label{fig:supervision}
\end{figure}

\textbf{Language-based policies can be trained with sparse natural language annotations:} Performance of \ourmethod trained with 10\% and 100\% natural language annotations is similar (and in both cases superior to \texttt{seq2seq} and \texttt{seq2seq2seq} trained on 100\% of data). Appendix \cref{fig:supervision} shows more detailed supervision curves.
Ablation experiments in \cref{tab:ablations} show that inference of latent training plans is important for this result: with no inference of latent instructions (i.e.\ training only on annotated demonstrations), performance drops from 56\% to 52\%. 
\cref{fig:plans} shows an example of the structure inferred for an unannotated trajectory: the model inserts reasonable segment boundaries and accurately labels each step.

\textbf{Language model pretraining improves automated decision-making.}
Ablation experiments in \cref{tab:ablations} provide details. Language model pretraining of $\poli$ and $\pole$ (on \emph{ungrounded} text) is crucial for good performance in the low-data regime: with 10\% of annotations, models trained from scratch complete 49\% of tasks (vs 56\% for pretrained models). 
We attribute this result in part to the fact that pretrained language models encode information about the common-sense structure of plans, e.g.\ the fact that \emph{slicing a tomato} first requires \emph{finding a knife}. Such models are well-positioned to adapt to ``planning'' problems that require modeling relations between natural language strings. These experiments point to a potentially broad role for pretrained language models in tasks that do not involve language as an input or an output. 

\begin{figure*}[t]
\centering
\includegraphics[width=\linewidth]{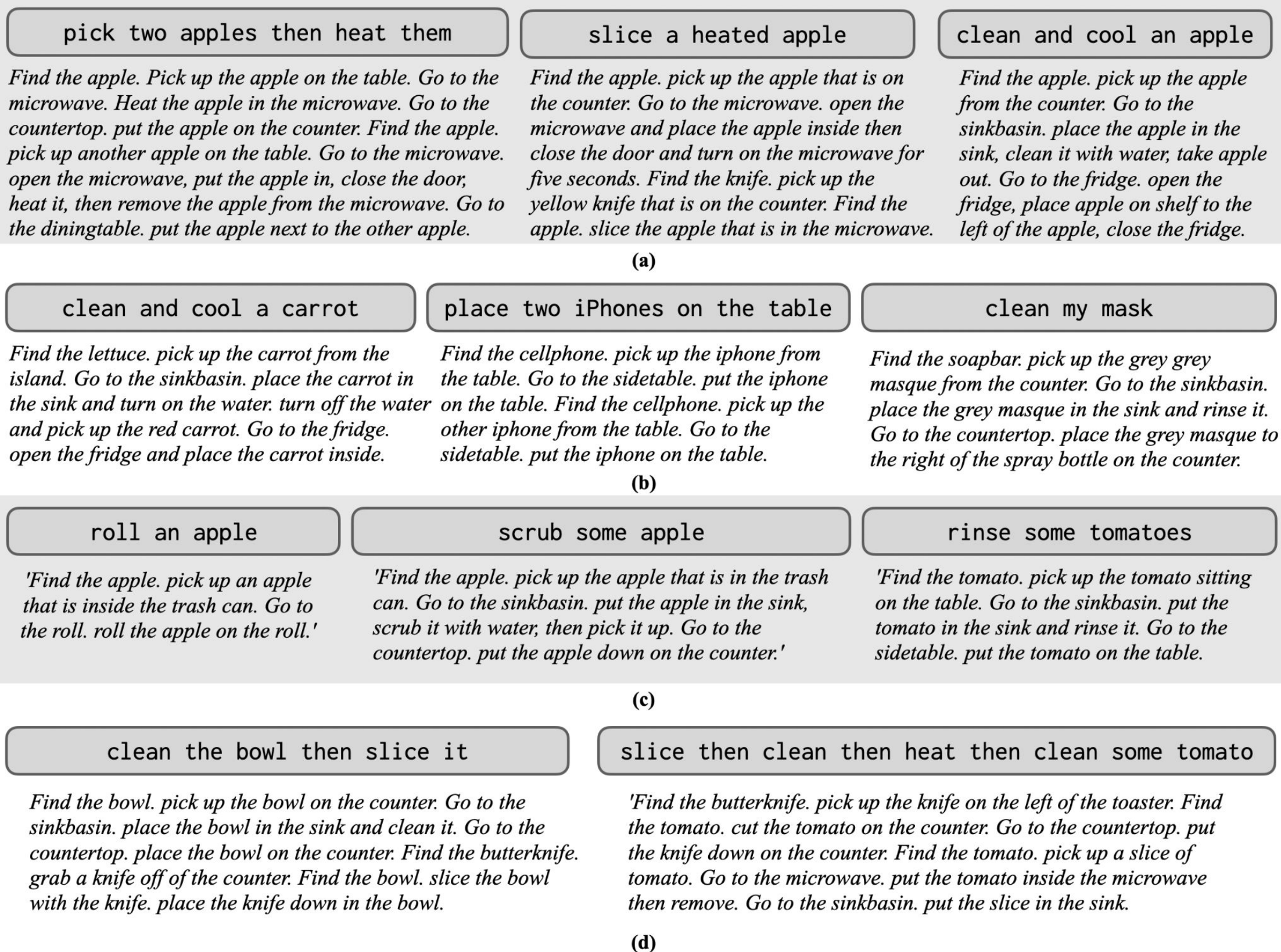}
\caption{Successes and failures of $\poli$ in out-of-distribution (OOD) settings including novel (a) sub-task orders (b) objects (c) verbs. The use of a pretrained LM as the backbone of the planning model means that models produce correct or plausible plans for many of these out-of-distribution goals. (d) Other failure modes: The model fails to predict actions based on the true affordances of objects and cannot generate arbitrarily long plans.
}
\label{fig:ood}
\end{figure*}

One especially interesting consequence of the use of language-based skills is our model's ability to produce high-level plans for \textbf{out-of-distribution goals}, featuring objects or actions that are not part of the ALFRED dataset at all. Examples are provided in \cref{fig:ood} and discussed in \cref{app:ood}. While additional modeling work is needed to generate low-level actions for these high-level plans, they point to \emph{generalization} as a key differentiator between latent language policies and ordinary hierarchical ones.

\section{Related Work}

Our approach draws on a large body of research at the intersection of natural language processing, representation learning, and autonomous control.

\textbf{Language-based supervision and representation} The use of natural language annotations to scaffold learning, especially in computer vision and program synthesis applications, has been the subject of a number of previous studies \cite{Branavan2009ReinforcementLF,Frome2013DeViSEAD,andreas-latent-lang,Wong2021LeveragingLT}. Here, we use language to support policy learning, specifically by using natural language instructions to discover compositional subtask abstractions that can support autonomous control.
Our approach is closely related to previous work on learning skill libraries from \emph{policy sketches} \cite{andreas-policy-sketch,Shiarlis2018TACOLT}; 
instead of the fixed skill inventory used by policy sketches, \ourmethod learns an open-ended, compositional library of behaviors indexed by natural language strings.

\textbf{Hierarchical policies} 
Hierarchical policy learning and temporal abstraction have been major areas of focus since the earliest research on reinforcement learning and imitation learning
\cite{McGovern,Konidaris,HierarchicalRE}. Past work typically relies on direct supervision or manual specification of the space of high-level skills \cite{options,Kulkarni2016HierarchicalDR} or fully unsupervised skill discovery \cite{maxq,optioncritic}. Our approach uses policy architectures from this literature, but aims to provide a mechanism for supervision that allows fine-grained control over the space of learned skills (as in fully supervised approaches) while requiring only small amounts of easy-to-gather human supervision.

\textbf{Language and interaction}
Outside of language-based supervision,
problems at the intersection of language and control include \emph{instruction following} \cite{mooney,Branavan2009ReinforcementLF,Tellex2011UnderstandingNL,anderson,Misra2017MappingIA}, \emph{embodied question answering} \cite{Das2018EmbodiedQA,Gordon2018IQAVQ} and dialog tasks \cite{Tellex2020RobotsTU}.
As in our work, representations of language learned from large text corpora facilitate grounded language learning \cite{alfworld}, and interaction with the environment can in turn improve the accuracy of language generation \cite{piglet}; future work might extend our framework for semi-supervised inference of plan descriptions to these settings as well.

\section{Conclusion}

We have presented \ourmethod, a framework for learning hierarchical policies from demonstrations sparsely annotated with natural language descriptions. Using these annotations, \ourmethod infers the latent structure of unannotated demonstrations, automatically segmenting them into subtasks and labeling each subtask with a compositional description. Learning yields a hierarchical policy in which natural language serves as an abstract representation of subgoals and plans: a \emph{controller} sub-policy maps from goals to natural language plan specifications, and a modular \emph{executor} that maps each component of the plan to a sequence of low-level actions. In simulated household environments, this model can complete abstract goals (like \emph{slice a tomato}) with accuracy comparable to state-of-the-art models trained and evaluated with fine-grained plans (\emph{find a knife}, \emph{carry the knife to the tomato}, \ldots).

While our evaluation has focused on household robotics tasks, the hierarchical structure inferred by \ourmethod is present in a variety of learning problems, including image understanding, program synthesis, and language generation. In all those domains, generalized versions of \ourmethod might offer a framework for building high-quality models using only a small amount of rich natural language supervision.

\section*{Acknowledgements}
We would like to thank Valts Blukis and Shikhar Murty for helpful discussions.
Also thanks to Joe O'Connor, Gabe Grand and the anonymous reviewers for their feedback on an early draft of the paper.

\bibliography{anthology,custom}

\begin{thebibliography}{54}
\expandafter\ifx\csname natexlab\endcsname\relax\def\natexlab#1{#1}\fi

\bibitem[{Anderson et~al.(2018)Anderson, Wu, Teney, Bruce, Johnson,
  S{\"u}nderhauf, Reid, Gould, and Hengel}]{anderson}
Peter Anderson, Qi~Wu, Damien Teney, Jake Bruce, Mark Johnson, Niko
  S{\"u}nderhauf, I.~Reid, Stephen Gould, and A.~V. Hengel. 2018.
\newblock Vision-and-language navigation: Interpreting visually-grounded
  navigation instructions in real environments.
\newblock \emph{2018 IEEE/CVF Conference on Computer Vision and Pattern
  Recognition}, pages 3674--3683.

\bibitem[{Andreas et~al.(2017)Andreas, Klein, and
  Levine}]{andreas-policy-sketch}
Jacob Andreas, D.~Klein, and Sergey Levine. 2017.
\newblock Modular multitask reinforcement learning with policy sketches.
\newblock \emph{International Conference of Machine Learning}.

\bibitem[{Andreas et~al.(2018)Andreas, Klein, and Levine}]{andreas-latent-lang}
Jacob Andreas, Dan Klein, and Sergey Levine. 2018.
\newblock Learning with latent language.
\newblock New Orleans, Louisiana. Association for Computational Linguistics.

\bibitem[{Bacon et~al.(2017)Bacon, Harb, and Precup}]{optioncritic}
P.~Bacon, Jean Harb, and Doina Precup. 2017.
\newblock The option-critic architecture.
\newblock In \emph{AAAI}.

\bibitem[{Baum et~al.(1970)Baum, Petrie, Soules, and Weiss}]{baum}
L.~Baum, T.~Petrie, George~W. Soules, and Norman Weiss. 1970.
\newblock A maximization technique occurring in the statistical analysis of
  probabilistic functions of markov chains.
\newblock \emph{Annals of Mathematical Statistics}, 41:164--171.

\bibitem[{Blukis et~al.(2021)Blukis, Paxton, Fox, Garg, and Artzi}]{blukis}
Valts Blukis, Chris Paxton, D.~Fox, Animesh Garg, and Yoav Artzi. 2021.
\newblock A persistent spatial semantic representation for high-level natural
  language instruction execution.
\newblock \emph{ArXiv}, abs/2107.05612.

\bibitem[{Branavan et~al.(2009)Branavan, Chen, Zettlemoyer, and
  Barzilay}]{Branavan2009ReinforcementLF}
S.~Branavan, Harr Chen, Luke Zettlemoyer, and R.~Barzilay. 2009.
\newblock Reinforcement learning for mapping instructions to actions.
\newblock In \emph{ACL}.

\bibitem[{Brockett(1993)}]{Brockett1993HybridMF}
R.~Brockett. 1993.
\newblock Hybrid models for motion control systems.

\bibitem[{Chen and Mooney(2011)}]{mooney}
David~L. Chen and R.~Mooney. 2011.
\newblock Learning to interpret natural language navigation instructions from
  observations.
\newblock In \emph{AAAI 2011}.

\bibitem[{Corona et~al.(2021)Corona, Fried, Devin, Klein, and
  Darrell}]{Corona2021ModularNF}
Rodolfo Corona, Daniel Fried, Coline Devin, D.~Klein, and Trevor Darrell. 2021.
\newblock Modular networks for compositional instruction following.
\newblock In \emph{NAACL}.

\bibitem[{Daniel et~al.(2012)Daniel, Neumann, and Peters}]{HierarchicalRE}
Christian Daniel, G.~Neumann, and Jan Peters. 2012.
\newblock Hierarchical relative entropy policy search.
\newblock \emph{J. Mach. Learn. Res.}, 17:93:1--93:50.

\bibitem[{Das et~al.(2018)Das, Datta, Gkioxari, Lee, Parikh, and
  Batra}]{Das2018EmbodiedQA}
Abhishek Das, Samyak Datta, Georgia Gkioxari, Stefan Lee, Devi Parikh, and
  Dhruv Batra. 2018.
\newblock Embodied question answering.
\newblock \emph{2018 IEEE/CVF Conference on Computer Vision and Pattern
  Recognition Workshops (CVPRW)}, pages 2135--213509.

\bibitem[{Dayan and Hinton(1992)}]{Dayan1992FeudalRL}
P.~Dayan and Geoffrey~E. Hinton. 1992.
\newblock Feudal reinforcement learning.
\newblock In \emph{NIPS}.

\bibitem[{Dietterich(1999)}]{maxq}
Thomas~G Dietterich. 1999.
\newblock \href {http://arxiv.org/abs/cs/9905014} {Hierarchical reinforcement
  learning with the {MAXQ} value function decomposition}.

\bibitem[{Fikes et~al.(1972)Fikes, Hart, and Nilsson}]{Fikes1972LearningAE}
R.~Fikes, P.~Hart, and N.~Nilsson. 1972.
\newblock Learning and executing generalized robot plans.
\newblock \emph{Artif. Intell.}, 3:251--288.

\bibitem[{Fox et~al.(2017)Fox, Krishnan, Stoica, and
  Goldberg}]{Fox2017MultiLevelDO}
Roy Fox, S.~Krishnan, I.~Stoica, and Ken Goldberg. 2017.
\newblock Multi-level discovery of deep options.
\newblock \emph{ArXiv}, abs/1703.08294.

\bibitem[{Frome et~al.(2013)Frome, Corrado, Shlens, Bengio, Dean, Ranzato, and
  Mikolov}]{Frome2013DeViSEAD}
Andrea Frome, G.~Corrado, Jonathon Shlens, Samy Bengio, J.~Dean, Marc'Aurelio
  Ranzato, and Tomas Mikolov. 2013.
\newblock Devise: A deep visual-semantic embedding model.
\newblock In \emph{NIPS}.

\bibitem[{Gordon et~al.(2018)Gordon, Kembhavi, Rastegari, Redmon, Fox, and
  Farhadi}]{Gordon2018IQAVQ}
Daniel Gordon, Aniruddha Kembhavi, Mohammad Rastegari, Joseph Redmon, D.~Fox,
  and Ali Farhadi. 2018.
\newblock Iqa: Visual question answering in interactive environments.
\newblock \emph{2018 IEEE/CVF Conference on Computer Vision and Pattern
  Recognition}, pages 4089--4098.

\bibitem[{He et~al.(2016)He, Zhang, Ren, and Sun}]{He2016DeepRL}
Kaiming He, X.~Zhang, Shaoqing Ren, and Jian Sun. 2016.
\newblock Deep residual learning for image recognition.
\newblock \emph{2016 IEEE Conference on Computer Vision and Pattern Recognition
  (CVPR)}, pages 770--778.

\bibitem[{Hoffman et~al.(2013)Hoffman, Blei, Wang, and
  Paisley}]{Hoffman2013StochasticVI}
M.~Hoffman, David~M. Blei, Chong Wang, and J.~Paisley. 2013.
\newblock Stochastic variational inference.
\newblock \emph{ArXiv}, abs/1206.7051.

\bibitem[{Hu et~al.(2019)Hu, Yarats, Gong, Tian, and
  Lewis}]{Hu2019HierarchicalDM}
Hengyuan Hu, Denis Yarats, Qucheng Gong, Yuandong Tian, and M.~Lewis. 2019.
\newblock Hierarchical decision making by generating and following natural
  language instructions.
\newblock In \emph{NeurIPS}.

\bibitem[{Jacob et~al.(2021)Jacob, Lewis, and
  Andreas}]{Jacob2021MultitaskingIS}
Athul~Paul Jacob, M.~Lewis, and Jacob Andreas. 2021.
\newblock Multitasking inhibits semantic drift.
\newblock \emph{ArXiv}, abs/2104.07219.

\bibitem[{Jiang et~al.(2019)Jiang, Gu, Murphy, and Finn}]{Jiang2019LanguageAA}
Yiding Jiang, S.~Gu, K.~Murphy, and Chelsea Finn. 2019.
\newblock Language as an abstraction for hierarchical deep reinforcement
  learning.
\newblock In \emph{NeurIPS}.

\bibitem[{Kaelbling and Lozano-P{\'e}rez(2011)}]{tamp}
L~P Kaelbling and T~Lozano-P{\'e}rez. 2011.
\newblock Hierarchical task and motion planning in the now.
\newblock \emph{2011 IEEE International}.

\bibitem[{Kim et~al.(2021)Kim, Bhambri, Singh, Mottaghi, and Choi}]{abp}
Byeonghwi Kim, Suvaansh Bhambri, Kunal~Pratap Singh, Roozbeh Mottaghi, and
  Jonghyun Choi. 2021.
\newblock Agent with the big picture: Perceiving surroundings for interactive
  instruction following.
\newblock In \emph{Embodied AI Workshop CVPR}.

\bibitem[{Kingma and Welling(2014)}]{Kingma2014AutoEncodingVB}
Diederik~P. Kingma and Max Welling. 2014.
\newblock Auto-encoding variational bayes.
\newblock \emph{CoRR}, abs/1312.6114.

\bibitem[{Konidaris et~al.(2012)Konidaris, Kuindersma, Grupen, and
  Barto}]{Konidaris}
G.~Konidaris, S.~Kuindersma, R.~Grupen, and A.~Barto. 2012.
\newblock Robot learning from demonstration by constructing skill trees.
\newblock \emph{The International Journal of Robotics Research}, 31:360 -- 375.

\bibitem[{Kulkarni et~al.(2016)Kulkarni, Narasimhan, Saeedi, and
  Tenenbaum}]{Kulkarni2016HierarchicalDR}
Tejas~D. Kulkarni, Karthik Narasimhan, A.~Saeedi, and J.~Tenenbaum. 2016.
\newblock Hierarchical deep reinforcement learning: Integrating temporal
  abstraction and intrinsic motivation.
\newblock In \emph{NIPS}.

\bibitem[{Loshchilov and Hutter(2019)}]{adamw}
I.~Loshchilov and F.~Hutter. 2019.
\newblock Decoupled weight decay regularization.
\newblock In \emph{ICLR}.

\bibitem[{Ma et~al.(2019)Ma, Lu, Wu, Al-Regib, Kira, Socher, and
  Xiong}]{Ma2019SelfMonitoringNA}
Chih-Yao Ma, Jiasen Lu, Zuxuan Wu, G.~Al-Regib, Z.~Kira, R.~Socher, and Caiming
  Xiong. 2019.
\newblock Self-monitoring navigation agent via auxiliary progress estimation.
\newblock \emph{ArXiv}, abs/1901.03035.

\bibitem[{McGovern and Barto(2001)}]{McGovern}
A.~McGovern and A.~Barto. 2001.
\newblock Automatic discovery of subgoals in reinforcement learning using
  diverse density.
\newblock In \emph{ICML}.

\bibitem[{Min et~al.(2021)Min, Chaplot, Ravikumar, Bisk, and
  Salakhutdinov}]{film}
So~Yeon Min, Devendra~Singh Chaplot, Pradeep Ravikumar, Yonatan Bisk, and
  Ruslan Salakhutdinov. 2021.
\newblock \href {http://arxiv.org/abs/2110.07342} {{FILM:} following
  instructions in language with modular methods}.
\newblock \emph{CoRR}, abs/2110.07342.

\bibitem[{Misra et~al.(2017)Misra, Langford, and Artzi}]{Misra2017MappingIA}
Dipendra~Kumar Misra, J.~Langford, and Yoav Artzi. 2017.
\newblock Mapping instructions and visual observations to actions with
  reinforcement learning.
\newblock In \emph{EMNLP}.

\bibitem[{Newell(1973)}]{Newell1973HumanPS}
A.~Newell. 1973.
\newblock Human problem solving.

\bibitem[{Pashevich et~al.(2021)Pashevich, Schmid, and Sun}]{et}
Alexander Pashevich, Cordelia Schmid, and Chen Sun. 2021.
\newblock \href {http://arxiv.org/abs/2105.06453} {Episodic transformer for
  vision-and-language navigation}.
\newblock \emph{CoRR}, abs/2105.06453.

\bibitem[{Rabiner(1989)}]{rabiner}
Lawrence~R. Rabiner. 1989.
\newblock A tutorial on hidden markov models and selected applications.
\newblock \emph{Proceedings of the IEEE}.

\bibitem[{Raffel et~al.(2020)Raffel, Shazeer, Roberts, Lee, Narang, Matena,
  Zhou, Li, and Liu}]{Raffel2020ExploringTL}
Colin Raffel, Noam~M. Shazeer, Adam Roberts, Katherine Lee, Sharan Narang,
  Michael Matena, Yanqi Zhou, W.~Li, and Peter~J. Liu. 2020.
\newblock Exploring the limits of transfer learning with a unified text-to-text
  transformer.
\newblock \emph{ArXiv}, abs/1910.10683.

\bibitem[{Sacerdoti(1973)}]{Sacerdoti1973PlanningIA}
E.~Sacerdoti. 1973.
\newblock Planning in a hierarchy of abstraction spaces.
\newblock \emph{Artif. Intell.}, 5:115--135.

\bibitem[{Shiarlis et~al.(2018)Shiarlis, Wulfmeier, Salter, Whiteson, and
  Posner}]{Shiarlis2018TACOLT}
K.~Shiarlis, Markus Wulfmeier, S.~Salter, S.~Whiteson, and I.~Posner. 2018.
\newblock Taco: Learning task decomposition via temporal alignment for control.
\newblock In \emph{ICML}.

\bibitem[{Shridhar et~al.(2020)Shridhar, Thomason, Gordon, Bisk, Han, Mottaghi,
  Zettlemoyer, and Fox}]{ALFRED20}
Mohit Shridhar, Jesse Thomason, Daniel Gordon, Yonatan Bisk, Winson Han,
  Roozbeh Mottaghi, Luke Zettlemoyer, and Dieter Fox. 2020.
\newblock \href {https://arxiv.org/abs/1912.01734} {{ALFRED: A Benchmark for
  Interpreting Grounded Instructions for Everyday Tasks}}.
\newblock In \emph{The IEEE Conference on Computer Vision and Pattern
  Recognition (CVPR)}.

\bibitem[{Shridhar et~al.(2021)Shridhar, Yuan, C{\^o}t{\'e}, Bisk, Trischler,
  and Hausknecht}]{alfworld}
Mohit Shridhar, Xingdi Yuan, Marc-Alexandre C{\^o}t{\'e}, Yonatan Bisk, Adam
  Trischler, and M.~Hausknecht. 2021.
\newblock Alfworld: Aligning text and embodied environments for interactive
  learning.
\newblock \emph{ArXiv}, abs/2010.03768.

\bibitem[{Shu et~al.(2018)Shu, Xiong, and Socher}]{Shu2018HierarchicalAI}
Tianmin Shu, Caiming Xiong, and R.~Socher. 2018.
\newblock Hierarchical and interpretable skill acquisition in multi-task
  reinforcement learning.
\newblock \emph{ArXiv}, abs/1712.07294.

\bibitem[{Singh et~al.(2020)Singh, Bhambri, Kim, Mottaghi, and
  Choi}]{singh2020moca}
Kunal~Pratap Singh, Suvaansh Bhambri, Byeonghwi Kim, Roozbeh Mottaghi, and
  Jonghyun Choi. 2020.
\newblock Moca: A modular object-centric approach for interactive instruction
  following.
\newblock \emph{arXiv preprint arXiv:2012.03208}.

\bibitem[{Suglia et~al.(2021)Suglia, Gao, Thomason, Thattai, and
  Sukhatme}]{embert}
Alessandro Suglia, Qiaozi Gao, Jesse Thomason, Govind Thattai, and Gaurav
  Sukhatme. 2021.
\newblock \href {https://arxiv.org/abs/2108.04927} {Embodied {BERT}: A
  transformer model for embodied, language-guided visual task completion}.
\newblock \emph{arXiv}.

\bibitem[{Sutton et~al.(1999)Sutton, Precup, and Singh}]{options}
R~S Sutton, D~Precup, and S~Singh. 1999.
\newblock Between {MDPs} and {semi-MDPs}: A framework for temporal abstraction
  in reinforcement learning.
\newblock \emph{Artif. Intell.}

\bibitem[{Tellex et~al.(2020)Tellex, Gopalan, Kress-Gazit, and
  Matuszek}]{Tellex2020RobotsTU}
Stefanie Tellex, N.~Gopalan, H.~Kress-Gazit, and Cynthia Matuszek. 2020.
\newblock Robots that use language.

\bibitem[{Tellex et~al.(2011)Tellex, Kollar, Dickerson, Walter, Banerjee,
  Teller, and Roy}]{Tellex2011UnderstandingNL}
Stefanie Tellex, T.~Kollar, Steven Dickerson, Matthew~R. Walter, A.~Banerjee,
  S.~Teller, and N.~Roy. 2011.
\newblock Understanding natural language commands for robotic navigation and
  mobile manipulation.
\newblock In \emph{AAAI}.

\bibitem[{Vaswani et~al.(2017)Vaswani, Shazeer, Parmar, Uszkoreit, Jones,
  Gomez, Kaiser, and Polosukhin}]{vaswani}
Ashish Vaswani, Noam Shazeer, Niki Parmar, Jakob Uszkoreit, Llion Jones,
  Aidan~N Gomez, Lukasz Kaiser, and Illia Polosukhin. 2017.
\newblock \href {http://arxiv.org/abs/1706.03762} {Attention is all you need}.

\bibitem[{Wainwright and Jordan(2008)}]{Wainwright2008GraphicalME}
Martin~J. Wainwright and M.I. Jordan. 2008.
\newblock Graphical models, exponential families, and variational inference.
\newblock \emph{Found. Trends Mach. Learn.}, 1:1--305.

\bibitem[{Wiseman et~al.(2018)Wiseman, Shieber, and
  Rush}]{Wiseman2018LearningNT}
Sam Wiseman, S.~Shieber, and Alexander~M. Rush. 2018.
\newblock Learning neural templates for text generation.
\newblock \emph{ArXiv}, abs/1808.10122.

\bibitem[{Wong et~al.(2021)Wong, Ellis, Tenenbaum, and
  Andreas}]{Wong2021LeveragingLT}
Catherine Wong, Kevin Ellis, J.~Tenenbaum, and Jacob Andreas. 2021.
\newblock Leveraging language to learn program abstractions and search
  heuristics.
\newblock In \emph{ICML}.

\bibitem[{Zellers et~al.(2021)Zellers, Holtzman, Peters, Mottaghi, Kembhavi,
  Farhadi, and Choi}]{piglet}
Rowan Zellers, Ari Holtzman, Matthew~E. Peters, R.~Mottaghi, Aniruddha
  Kembhavi, Ali Farhadi, and Yejin Choi. 2021.
\newblock Piglet: Language grounding through neuro-symbolic interaction in a 3d
  world.
\newblock In \emph{ACL/IJCNLP}.

\bibitem[{Zhang and Chai(2021)}]{hitut}
Yichi Zhang and Joyce Chai. 2021.
\newblock \href {http://arxiv.org/abs/2106.03427} {Hierarchical task learning
  from language instructions with unified transformers and self-monitoring}.
\newblock \emph{CoRR}, abs/2106.03427.

\bibitem[{Ziebart et~al.(2013)Ziebart, Bagnell, and Dey}]{ziebart2013principle}
Brian~D Ziebart, J~Andrew Bagnell, and Anind~K Dey. 2013.
\newblock The principle of maximum causal entropy for estimating interacting
  processes.
\newblock \emph{IEEE Transactions on Information Theory}, 59(4):1966--1980.

\end{thebibliography}
\bibliographystyle{acl_natbib}

\newpage
\appendix
\label{appendix}

\section{Out-of-distribution Generalization}
\label{app:ood}
One of the advantages of language-based skill representations over categorical representations is open-endedness: \ourmethod does not require pre-specification of a fixed inventory of goals or actions. As a simple demonstration of this potential for extensibility, we design goal prompts consisting of novel object names, verbs and skill combinations not seen at training time, and test the model's ability to generalize to out-of-distribution samples across the three categories. Some roll-outs can be seen in \cref{fig:ood}. We observe the following:

\paragraph{Novel sub-task combinations} We qualitatively evaluate the ability of the model to generalize systematically to novel subtask combinations and subtask ordering not encountered at training time. Examples are shown in \cref{fig:ood}. For example, we present the model with the goal  \emph{slice a heated apple}; in the training corpus, objects are only heated \emph{after} being sliced. It can be seen in \cref{fig:ood} that the model able correctly orders the two subtasks.
The model additionally generalizes to new combinations of tasks such as \emph{clean and cool an apple}.

\paragraph{Novel objects and verbs} The trained model also exhibits some success at generalizing novel object categories such as \emph{carrot} and \emph{mask}. In the carrot example, an incorrect \emph{Find the lettuce} example is generated at the first step, but subsequent subtasks refer to a carrot (and apply the correct actions to it). The model also generalizes to new but related verbs such as \emph{scrub} but fails at ones like \emph{squash} that are unrelated to training goals.

\paragraph{Limitations} One shortcoming of this approach is that affordances and constraints are incompletely modeled. Given a (physically unrealizable) goal \emph{clean the bowl and then slice it}, the model cannot detect the impossible goal and instead generates a plan involving slicing the bowl. Another shortcoming of the model is the ability to generalize to goals that may involve considerably larger number of subgoals than goals seen at training time. For plans that involve very long sequences of skills (\emph{slice then clean then heat\ldots}) the generated plan skips some subtasks \cref{fig:ood}.

\begin{figure*}
\centering
\includegraphics[width=0.9\linewidth]{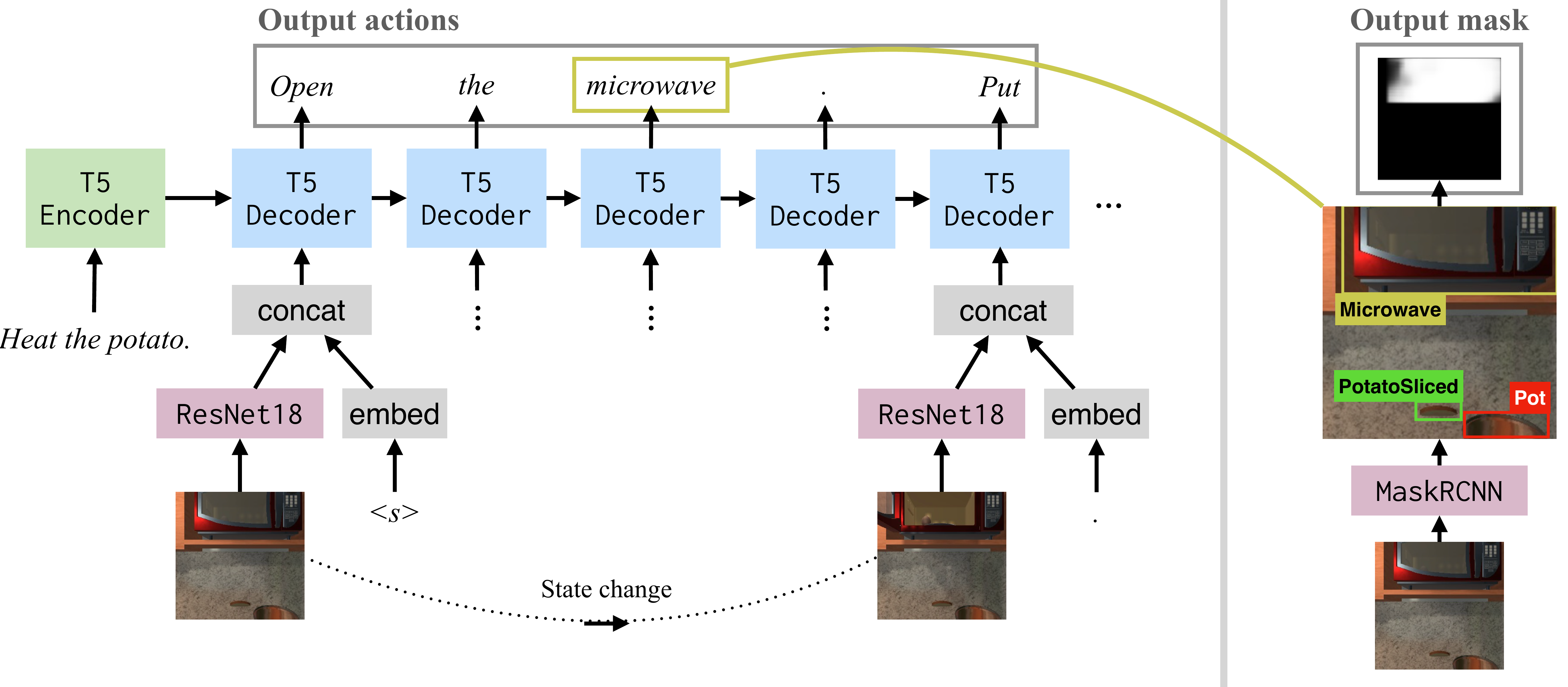}
\caption{Model architecture for $\pole$, seq2seq and seq2seq2seq: Language parametrized sub-task/goal is input to the encoder and actions templated in natural language are generated sequentially token-wise. The predictions made are conditioned on the visual field of view of the agent at every time step along with the token generated the previous time step. At the end of every low-level action (when '.' is generated) the action the executed. For manipulation actions, the mask corresponding to the the object predicted is selected from the predictions of a MaskRCNN model on the visual state. Navigation actions do not operate over objects. Once the action is taken, the environment returns the updated visual state and the policy continues to be unrolled until termination ($\actstop$).
}
\label{fig:architecture}
\end{figure*}

\section{Initialization: Segmentation Step}
\label{app:init}
The training data contains no \actstop actions, so $\pole$ cannot be initialized by training on $\datasetann$. Using a randomly initialized $\pole$ during the segmentation step results in extremely low-quality segmentations. Instead, we obtain an initial set of segmentations via \emph{unsupervised} learning on low-level action sequences. 

In particular, we obtain initial segmentations using the Baum--Welch algorithm for unsupervised estimation of hidden Markov models \citep{baum}. We replace string-valued latent variables produced by $\poli$ with a discrete set of hidden states (in our experiments, we found that three hidden states sufficed). Transition and emission distributions, along with maximum \emph{a posteriori} sequence labels, are obtained by running the expectation--maximization algorithm on state sequences. We then insert segment boundaries (and an implicit \actstop action) at every transition between two distinct hidden states.
Evaluated against ground-truth segmentations from the ALFRED training set, this produces an action-level accuracy of 87.9\%. %
The detailed algorithm can be found in \citet{baum}.

\section{Model Architecture: Details}
\label{app:model}
The controller policy $\poli$ is a fine-tuned T5-small model. The executor policy $\pole$ decodes the low-level sequence of actions conditioned on the first-person visual observations of the agent. We use the same architecture across the remaining baselines too. \cref{fig:architecture} depicts the architecture of the image-conditioned T5 model. %
In addition to task specifications, we convert low-level actions to templated commands: for example, \texttt{put(cup,table)} becomes \emph{put the cup on the table}. %
These are parsed to select actions to send to the ALFRED simulator. During training, both models are optimized using the AdamW algorithm \cite{adamw} with a learning rate of 1e-4, weight decay of 0.01, and $\epsilon = $ 1e-8. 
We use a MaskRCNN model to generate action masks, selecting the predicted mask labeled with the class of the object name generated by the action decoder. The same model architecture is used across all baselines.

\section{Role of trajectory length}
\label{app:hierarchy}
We conduct an additional set of ablation experiments aimed at clarifying what aspects of the demonstrated trajectories \ourmethod is better able to model than baselines. We begin by observing that most actions in our data are associated with navigation, with sequences of object manipulation actions (like those depicted in \cref{fig:plans}) constituting only about 20\% of each trajectory. We construct an alternative version of the dataset in which all navigation subtasks are replaced with a single \texttt{TeleportTo} action. 
This modification reduces average trajectory length from 50 actions to 9. 
In this case \ourmethod and \texttt{seq2seq2seq} perform comparably well (55.6\% success rate and 56.7\% success rate respectively), and only slightly better than \texttt{seq2seq} (53.6\% success rate). Thus, while \ourmethod (and all baselines) perform quite poorly at navigation skills, \emph{identifying} these skills and modeling their conditional independence from other trajectory components seems to be crucial for effective learning of other skills in the long-horizon setting. Hierarchical policies are still useful for modeling these shorter plans, but by a smaller margin than for long demonstrations.

\begin{figure}[t]
\centering
\includegraphics[width=\linewidth]{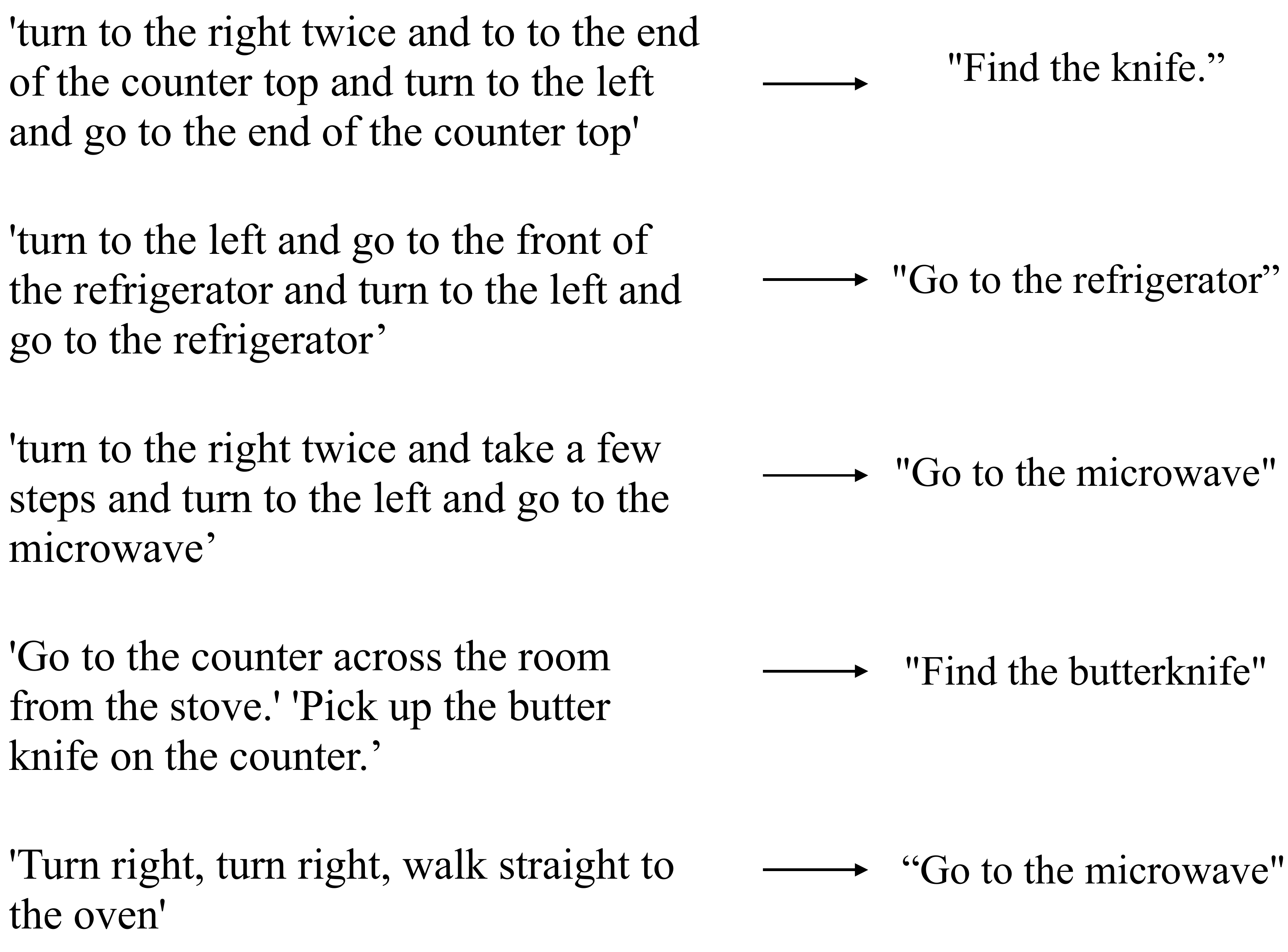}
\caption{Modified navigation annotations. Navigation instructions are converted to simpler object/location-oriented navigation goals using by creating templated plans from ALFRED dataset metadata.
}
\label{fig:nav-annotations-modified}
\end{figure}

\begin{table*}[t]
\centering
\begin{small}
\begin{tabular}{@{}lcccccccc@{}}
\toprule
\multirow{2}{*}{\textbf{Method}} & \multicolumn{5}{c}{\textbf{Training time}} && \multicolumn{2}{c}{\textbf{Inference time}}\\
& {\bf Goal} & {\bf Instructions} & {\bf Program} & {\bf Alignments} & {\bf Depth} &&{\bf Goal} & {\bf Instructions}\\
\cmidrule{2-6} \cmidrule{8-9}
\ourmethod & \cmark & 10\% & \xmark & \xmark & \xmark && \cmark & \xmark \\
seq2seq & \cmark & \xmark & \xmark& \xmark & \xmark&& \cmark &\xmark \\
seq2seq2seq  & \cmark & \cmark & \xmark & \xmark & \xmark && \cmark & \xmark\\

S+\cite{ALFRED20}  & \cmark & \cmark & \xmark & \cmark & \xmark && \cmark & \cmark \\
MOCA\cite{singh2020moca} & \cmark & \cmark & \xmark & \cmark & \xmark && \cmark & \cmark \\
Modular \cite{Corona2021ModularNF}& \cmark & \cmark & \cmark & \cmark & \xmark && \cmark & \cmark \\
ABP \cite{abp}& \cmark & \cmark & \xmark & \cmark & \xmark && \cmark & \cmark \\
EmBERT \cite{embert}& \cmark & \cmark & \xmark & \cmark & \xmark && \cmark & \cmark \\
ET \cite{et}& \cmark & \cmark & \xmark & \cmark & \xmark && \cmark & \cmark \\
HLSM\cite{blukis} & \cmark & \xmark & * & * & \cmark && \cmark & \xmark \\
HiTUT \cite{hitut}& \cmark & \xmark & \cmark & \cmark & \cmark && \cmark & \xmark \\
FILM \cite{film}& \cmark & \xmark & \cmark & \cmark & \cmark && \cmark & \xmark \\
\bottomrule
\end{tabular}
\label{fig:dataset_comparison}
\end{small}
\caption{Detailed comparison of information available to models and baselines at training time and inference. {\small *Re-derived using a rule-based segmentation procedure}} 
\label{tab:info-diff}
\end{table*}

\section{Navigation Instructions}
\label{app:nav}
The original ALFRED dataset contains detailed instructions for navigation collected post-hoc after the demonstrations are generated. For example, the sub-task specification associated with finding an apple might be given as \emph{Go straight and turn to the right of the fridge and take a few steps ahead and look down}. Such instructions cannot be used for high-level planning, as they can only be generated with advance knowledge of the environment layout; successful behavior in novel environments requires  exploration or explicit access to the environment’s map. 

To address the mismatch between the agent's knowledge and the information needed to generate detailed navigation instructions, we navigation instructions in the ALFRED dataset with templated instructions of the form \emph{Go to the \emph{[object]}} (for appliances and containers)  and \emph{Find the \emph{[object]}} (for movable objects). Because the ALFRED dataset provides PDDL plans for each demonstration, we can obtain the name of the target [object] directly from these plans. Examples are shown in \cref{fig:nav-annotations-modified}.

\end{document}